\documentclass{article}
\usepackage{ijcai22}

\usepackage{times}
\usepackage{soul}
\usepackage{url}
\usepackage[hidelinks]{hyperref}
\usepackage[utf8]{inputenc}
\usepackage[small]{caption}
\usepackage{graphicx}
\usepackage{amsmath}
\usepackage{amsthm}
\usepackage{booktabs}
\usepackage{algorithm}
\usepackage{algorithmic}
\usepackage{amssymb}
\newcommand{\model}{MaKEr}

\usepackage{xcolor}

\title{Meta-Learning Based Knowledge Extrapolation for Knowledge Graphs in the Federated Setting}

\author{
Mingyang Chen$^1$\footnote{Equal Contribution.} \and
Wen Zhang$^2$\footnotemark[1] \and
Zhen Yao$^2$\and
Xiangnan Chen$^2$\and
Mengxiao Ding$^3$\and \\
Fei Huang$^4$\And
Huajun Chen$^{1,5,6}$\footnote{Corresponding author.}
\affiliations
$^1$College of Computer Science and Technology, Zhejiang University\\
$^2$School of Software Technology, Zhejiang University\\
$^3$Huawei Technologies Co., Ltd. 
$^4$Alibaba Group \\
$^5$ZJU-Hangzhou Global Scientific and Technological Innovation Center \\
$^6$Alibaba-Zhejiang University Joint Institute of Frontier Technologies
\emails
\{mingyangchen, zhang.wen, yz0204, xnchen2020, huajunsir\}@zju.edu.cn,  \\
dingmengxiao@huawei.com,
f.huang@alibaba-inc.com
}

\begin{document}

\maketitle

\begin{abstract}
We study the knowledge extrapolation problem to embed new components (i.e., entities and relations) that come with emerging knowledge graphs (KGs) in the federated setting. In this problem, a model trained on an existing KG needs to embed an emerging KG with unseen entities and relations. 
To solve this problem, we introduce the meta-learning setting, where a set of tasks are sampled on the existing KG to mimic the link prediction task on the emerging KG. 
Based on sampled tasks, we meta-train a graph neural network framework that can construct features for unseen components based on structural information and output embeddings for them.
Experimental results show that our proposed method can effectively embed unseen components and outperforms models that consider inductive settings for KGs and baselines that directly use conventional KG embedding methods\footnote{Source code is available at \url{https://github.com/zjukg/MaKEr}.}.
\end{abstract}

\section{Introduction}


Knowledge graphs (KGs) are expressive data structures that consist of a large number of triples with the form of \textit{(head entity, relation, tail entity)}. 
Nowadays, many large-scale KGs have become essential data supports for an increasing number of applications~\cite{recsys,QA-GNN}.
With the development of KGs, they are no longer only applied in a centralized way where all the triples of a KG can be accessed on one device, but also in a decentralized manner. 
To our best knowledge, many mobile applications build personal KGs on users' devices, and naturally, new KGs on new devices are emerging at any given moment. 
However, conventional large-scale KGs are well known to be incomplete; thus, emerging KGs also suffer from this problem.
For KG completion, extensive research has been devoted to predicting missing links by learning low-dimensional vector representations (a.k.a, knowledge graph embeddings) for entities and relations that proved effective.

\begin{figure}[t]
\centering
\includegraphics[scale=0.6]{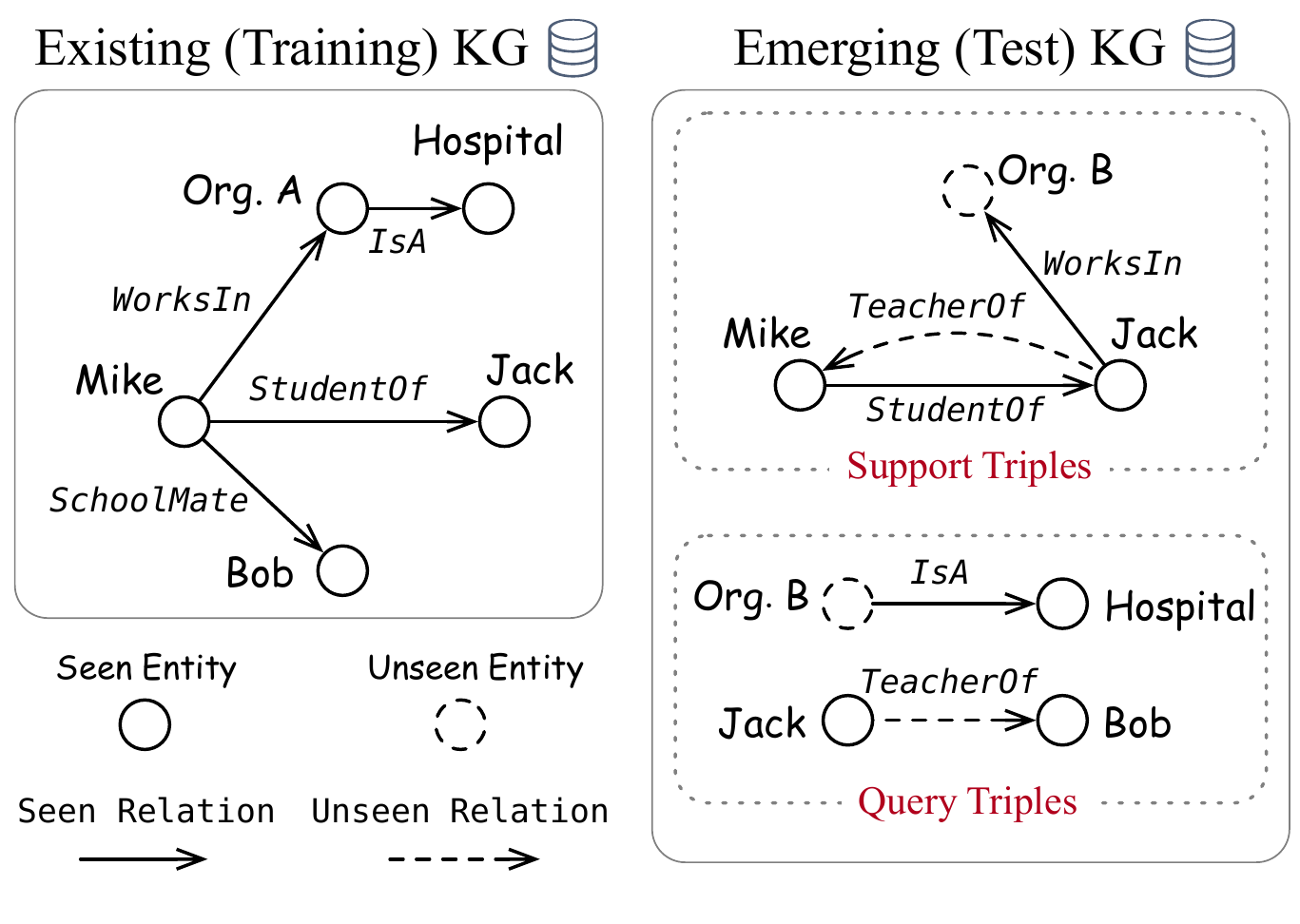}
\caption{The challenges for embedding emerging KGs in the federated setting.}
\label{fig:intro}
\end{figure}

Nevertheless, knowledge graph embedding (KGE) methods are not perfect in real-world applications.
Conventional KGE methods cannot deal with new relations and entities since they learn embeddings for a fixed set of entities and relations. While as mentioned above, an emerging KG is often accompanied by new relations and entities.
As shown in Fig.~\ref{fig:intro}, the emerging KG contains an unseen entity \texttt{Org.B} and an unseen relation \texttt{TeacherOf}, so a KGE model trained on the existing KG cannot be applied to query missing triples about such unseen components based on support triples in the emerging KG.

Although existing methods \cite{GraIL} focusing on inductive settings for KG completion can handle unseen entities in a new KG, they cannot consider both unseen relations and entities. 
Moreover, such inductive KG completion methods cannot take advantage of seen entities since no transferable information of seen entities, like trained vector representations, is considered; unless the existing KG is integrated with the emerging KG.
However, such KG integration may not be allowed for multiple reasons in real applications, such as data privacy. We call this scenario KGs in the federated setting~\cite{FedAvg}, namely, these KGs can use the same model while their data are not
shared explicitly.

Motivated by the challenges faced with this real-world scenario, we raise a research question: \textit{Can we have an embedding-based model trained on seen relations and entities and generalize to unseen relations and entities for KGs in the federated setting?} 

Formally, the link prediction task for an emerging KG can be viewed as predicting the plausibility of a set of query triples based on another set of support triples, as shown in Fig.~\ref{fig:intro}. 
To handle the unseen components in the emerging KG, inspired by the ability of ``learning to learn" brought by meta-learning \cite{mann}, we formulate a set of tasks consisting of support triples and query triples on the existing KG to mimic the link prediction task in the emerging KG, and learn to embed unseen components in each task. Furthermore, we propose a Graph Neural Network (GNN) framework that can embed both seen and unseen components based on support triples of a task.
Our model mainly contains three modules: 
1) a relation feature representation module, in which we construct a Relation Position Graph (RPG) from triples to reveal relative positions between relations and construct relation features based on it;
2) an entity feature representation module where we use the connections between entities and their neighbors to represent their features;
3) a GNN that encodes features and outputs embeddings for both seen and unseen components to achieve knowledge extrapolation.

This model is meta-trained on tasks sampled from the existing KG and learns the ability to embed unseen components for a task; thus, the learned model can generalize to an emerging KG with unseen components. 
To evaluate our method, we introduce datasets for knowledge extrapolation in the federated setting, which are sampled from KG benchmarks.
The evaluation results show that our method outperforms models that only consider unseen entities under inductive settings for KGs and baselines that calculate embeddings for unseen components based on conventional KGE assumptions.

\section{Related Work}

\paragraph{Knowledge Graph Embedding.}
Lots of KGE models~\cite{kgembedding} have been proposed to embed KGs into low-dimensional vector spaces.
Different methods map entities and relations into different vector spaces and design various score functions.
TransE \cite{TransE} is a representative method that maps entities and relations as vectors in the same real space, and relations are interpreted as translation vectors between entities for triples. 
Moreover, ComplEx \cite{ComplEx} and RotatE \cite{RotatE} embed KGs into complex spaces and propose more sophisticated score functions. However, conventional KGE models can only embed a fixed set of components and cannot generalize to unseen entities and relations after model training.

\paragraph{Inductive Settings for Knowledge Graph.}
Some existing works have proposed models applicable in inductive settings where there are unseen entities during the test.
\cite{OOKB} and \cite{LAN} learn to embed unseen entities by neighborhood aggregation based on GNNs, while they can only embed unseen entities connected with the training KG. \cite{GraIL}, \cite{TACT} and \cite{INDIGO} train KG completion methods and can generalize to a new KG with unseen entities. However, they cannot generalize to unseen relations and cannot take advantage of seen entities during the test.

\paragraph{Meta-Learning.}
Meta-learning, known as ``learning to learn", focuses on learning to generalize over the distribution of tasks but not data points.
Metric-based methods \cite{prototypical,matching} learn generalizable parameterized metrics from a set of training tasks. 
Optimized-based methods \cite{maml} learn the optimization of model parameters given the gradients on a task.
Black-box methods \cite{mann} train a model that can represent another model's parameters by standard supervised learning.
Some existing works~\cite{GMatching,MetaR} apply meta-learning on KGs to solve the few-shot problem, but not the problem in this paper.

\paragraph{Knowledge Graph and GNN.} Graph neural networks \cite{GCN} encode representations for nodes in graphs via their neighbor structures. Recently, R-GCN \cite{RGCN} considers relation-specific transformations for neighbor aggregation of a node.
CompGCN \cite{CompGCN} leverages entity-relation composition operations to embed entities and relations jointly and generalizes several previous multi-relational GCNs.
We extend CompGCN as the GNN for updating entity and relation features in our model by replacing composition operators with a linear transformation for entity-relation aggregations. Such linear transformations are more general and flexible when dimensions for entities and relations are diverse.

\paragraph{Federated Settings for Knowledge Graph.}
Traditional research for knowledge graphs focuses on centralized settings, while KGs are evolving and may be built by different controllers \cite{FedE}. Some existing works \cite{FedSage} train a generalizable graph mining model without sharing graph data from multiple local systems. \cite{FKGE} tries to embed components from different KGs while being privacy-preserving. 
However, previous works learn embeddings for KGs from different data sources based on federated learning \cite{FedAvg}, and all the entities and relations for test triples are seen during training. However, the problem in our work focuses on the emerging KG with unseen components in the federated setting.

\section{Problem Formulation}
\label{sec:problem-formulation}

A knowledge graph is defined as $\mathcal{G} = (\mathcal{E}, \mathcal{R}, \mathcal{T})$, where $\mathcal{E}$ denotes a set of entities, $\mathcal{R}$ is a set of relations, and $\mathcal{T}$ is a set of triples. Specifically, $\mathcal{T}=\{(h,r,t)\} \subseteq \mathcal{E} \times \mathcal{R} \times \mathcal{E}$, where $h, t \in \mathcal{E}$ and $r \in \mathcal{R}$. 
The link prediction task for $\mathcal{G}$ refers to the problem of predicting $e \in \mathcal{E}$, given an entity and a relation, namely, $(h, r, ?)$ or $(?, r, t)$, to make $(h, r, e)$ or $(e, r, t)$ a new true triple for completing $\mathcal{G}$.

In the context of traditional knowledge graph completion \cite{TransE}, to evaluate the link prediction ability of a model, there are two part of triples, training (a.k.a, support) triples $\mathcal{T}_{sup}$ and testing (a.k.a, query) triples $\mathcal{T}_{que}$.
Specifically, $\mathcal{T}_{sup}$ is used for training a model $\mathcal{M}$ which usually has the ability to score the plausibility of a triple.
Furthermore, $\mathcal{T}_{que}$ are ground-truth triples for evaluating the learned model. 
For example, for tail link prediction, given an ground-truth triple $(h,r,t) \in \mathcal{T}_{que}$, we rank $\mathcal{M}(h,r,t)$ over its candidates triples $\{\mathcal{M}(h,r,e) | e \in \mathcal{E}, (h,r,e) \notin \mathcal{T}_{sup} \cup  \mathcal{T}_{que}\}$. The higher the ranks for ground-truth triples in $\mathcal{T}_{que}$, the more effective the model is.

Next, based on the above definitions, we formally define the problem of knowledge extrapolation for knowledge graphs in the federated setting. Given a training KG $\mathcal{G}^{tr} = (\mathcal{E}^{tr}, \mathcal{R}^{tr}, \mathcal{T}^{tr})$, we aim at training a triple scoring model $\mathcal{M}$ that can generalize to a test KG $\mathcal{G}^{te} = (\mathcal{E}^{te}, \mathcal{R}^{te}, \mathcal{T}^{te}_{sup}, \mathcal{T}^{te}_{que})$, where $\mathcal{E}^{tr} \neq \mathcal{E}^{te}$, $\mathcal{E}^{tr} \cap \mathcal{E}^{te} \neq \emptyset$ and $\mathcal{R}^{tr} \neq \mathcal{R}^{te}$, $\mathcal{R}^{tr} \cap \mathcal{R}^{te} \neq \emptyset$. Moreover, this problem has two constrains: 1) \textit{knowledge extrapolation}, compared with the traditional knowledge graph completion task, the model training is only conducted on the $\mathcal{T}^{tr}$ but not on the $\mathcal{T}^{te}_{sup}$, and triples in $\mathcal{T}^{te}_{sup}$ are only used for revealing the connections of entities and relations in $\mathcal{E}^{te}$ and $\mathcal{R}^{te}$; 2) \textit{federated setting}, $\mathcal{G}^{tr}$ and $\mathcal{G}^{te}$ are not allowed to be integrated together.

\section{Methodology}
\subsection{Meta-Learning Setting}
\label{sec:meta-learning-setting}


The main problem here is how to embed unseen entities and relations on $\mathcal{G}^{te}$ with effective representations that can handle KG completion.
Inspired by the concept of ``learning to learn" from meta-learning, with the goal of ``embedding unseen entities and relations" on the test KG, we can formulate a set of tasks on the training KG with simulated unseen entities and relations to mimic the test environment, and learn a model on such tasks to achieve ``learning to embed unseen entities and relations". 

Specifically, each task $\mathcal{S}^{i} = (\mathcal{E}^{i}, \mathcal{R}^{i}, \mathcal{T}^{i}_{sup}, \mathcal{T}^{i}_{que})$ over a task distribution $p(\mathcal{S})$ corresponds to a sub-KG sampled from $\mathcal{G}^{tr}$. Although $\mathcal{E}^{i}$ and $\mathcal{R}^{i}$ are sampled from $\mathcal{E}^{tr}$ and $\mathcal{R}^{tr}$, we re-label a part of entities and relations and treat them as unseen entities and relations. A task $\mathcal{S}^{i}$ is defined as follows:
\begin{equation}
\begin{aligned}
    \mathcal{S}^{i} &= \left(
    \mathcal{E}^{i} = (\mathcal{\widehat{E}}^{i}, \mathcal{\widetilde{E}}^{i}), 
    \mathcal{R}^{i} = (\mathcal{\widehat{R}}^{i}, \mathcal{\widetilde{R}}^{i}), 
    \mathcal{T}^{i}_{sup}, 
    \mathcal{T}^{i}_{que}\right),
\label{eq:meta-train-task}
\end{aligned}
\end{equation}
where 
$\mathcal{\widehat{E}}^{i} \in \mathcal{E}^{tr}$ are seen entities and  
$\mathcal{\widetilde{E}}^{i} \notin \mathcal{E}^{tr}$ are unseen entities;
$\mathcal{\widehat{R}}^{i} \in \mathcal{R}^{tr}$ 
and 
$\mathcal{\widetilde{R}}^{i} \notin \mathcal{R}^{tr}$.
Based on the tasks sampled from $\mathcal{G}^{tr}$ for meta-learning, the meta-training objective is learning to embed both seen and unseen entities and relations based on support triples, to maximize the score of query triples as follows: 
\begin{equation}
    \max_{\theta} \mathbb{E}_{\mathcal{S}^{i} \sim p(\mathcal{S})} \left[ \sum_{(h,r,t) \in \mathcal{T}^{i}_{que}}
    \frac{1}{|\mathcal{T}^{i}_{que}|}
    \mathcal{M}_{\theta}(h,r,t|\mathcal{T}^{i}_{sup}) \right],
\end{equation}
where $\mathcal{M}$ is an arbitrary model that can calculate plausibility scores for query triples based on support triples. 

However, we don't have an existing model $\mathcal{M}$ which fits this problem. Hence, we design a model that can embed both seen and unseen entities and relations based on support triples of each sampled task, and we describe the details of the proposed model in the following.



\begin{figure*}[t]
\centering
\includegraphics[scale=0.6]{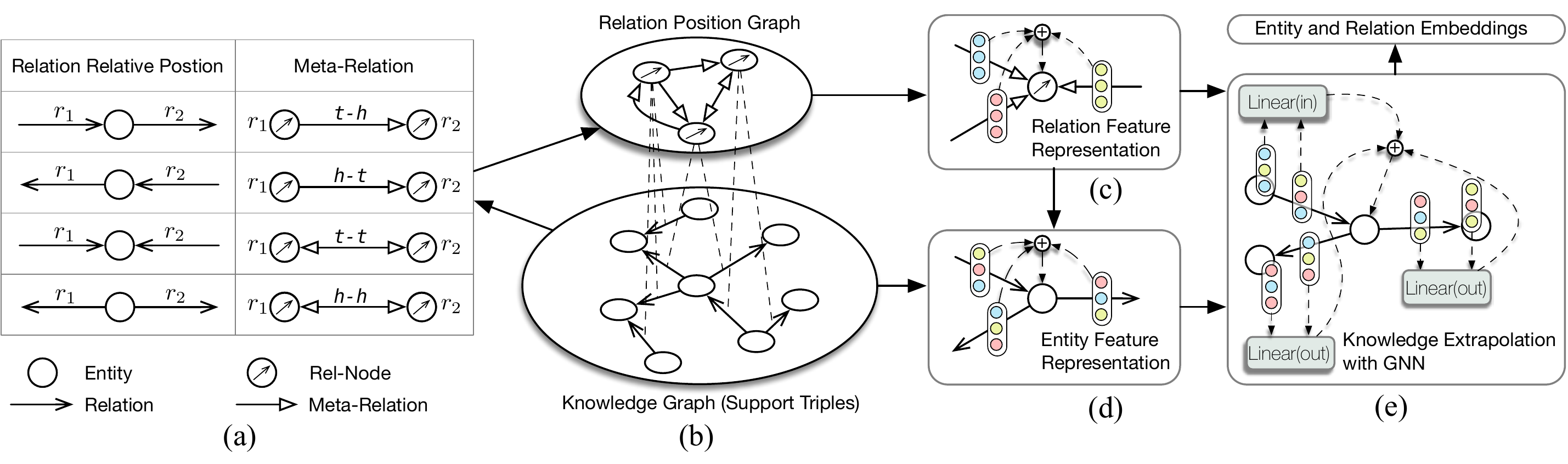}
\caption{Overview of our proposed model. (a)(b) Construct RPG from the support triples of a task; (c)(d)(e) Three modules of the proposed model that represent features for entities and relations and output embeddings for them.}
\label{fig:model}
\end{figure*}

\subsection{Proposed Model}

The overall of our model is a Graph Neural Network (GNN) framework and follows the encoder-decoder paradigm as previous works that apply GNN on KGs \cite{RGCN}.
Generally, the encoder is a GNN structure that takes feature representations as input and outputs the embeddings; the decoder is a KGE method that takes embeddings of the components of a triple as input and outputs the score.
Note that we do not use any entity or relation attributes to get the feature representation since our model only considers structure information of KGs, which is a common scenario and makes our model can be generalized to more applications.

Naturally, for a task $\mathcal{S}^{i}$, the feature representations of seen entities $e \in \mathcal{\widehat{E}}^{i}$ and relations $r \in \mathcal{\widehat{R}}^{i}$ can be looked up from an entity feature matrix $\mathbf{E}^{tr} \in \mathbb{R}^{|\mathcal{E}^{tr}| \times d_{e}}$ and a relation feature matrix $\mathbf{R}^{tr} \in \mathbb{R}^{|\mathcal{R}^{tr}| \times d_{r}}$, which can be randomly initialized based on the training KG $\mathcal{G}^{tr}$ and are learned during training. 
However, the difficulty is how to represent unseen entities and relations effectively. 
To solve this problem, we design the following three modules.
The overview of our model can be found in Fig.~\ref{fig:model}, and we describe it based on a specific sampled task $\mathcal{S}^{i}$ as illustrated in Equ.~(\ref{eq:meta-train-task}).

\subsubsection{Relation Feature Representation}

In the literature about transfer learning of GNNs \cite{EGI}, node degrees can be used as node features since they are sensitive to changes in the graph structures. In our work, we also try to find some structure-respecting features to represent relations.
To represent a relation as intuitional as a node in graphs, we construct a Relation Position Graph from the original knowledge graph (i.e., support triples $\mathcal{T}_{sup}^{i}$ for a task $\mathcal{S}^{i}$), where each node in the RPG represents one type of relation in the KG, as shown in Fig.~\ref{fig:model}(b).

We define four relationships between nodes in RPG, based on four relation relative positions (Fig.~\ref{fig:model}(a)). 
To avoid confusion, we refer to such ``relationships between relations'' as \textit{meta-relations} and refer to nodes that represent relations as \textit{rel-nodes}. 
For instance, ($r_1$, \texttt{t-h}, $r_2$) denotes that a \underline{t}ail entity linked to $r_1$ is the \underline{h}ead entity linked to $r_2$. 
For two specific relations, if one type of relative position for them appears in the support triples, their corresponding rel-nodes in RPG will be connected by the corresponding meta-relation. 
After building the RPG, as shown in Fig.~\ref{fig:model}(c), we represent the feature of an unseen relation $r$ by its neighbor meta-relations in RPG:
\begin{equation}
    \mathbf{h}_{r} = \frac{1}{|\mathcal{N}(r)|} \sum_{m \in \mathcal{N}(r)} \mathbf{h}_{m},
\label{eq:rel-repr}
\end{equation}
where $\mathcal{N}(r)$ denotes the set of in-going meta-relations of the rel-node representing $r$. Note that we only consider in-going meta-relations here, since \texttt{t-h} is the inverse of \texttt{h-t}, and \texttt{h-h} and \texttt{t-t} are double-direction, thus out-going meta-relations can be indicated by in-going meta-relations.
Moreover, $\mathbf{h}_{m} \in \mathcal{H}_{M}$ is the vector representation of a specific meta-relation, and $\mathcal{H}_{M} = (\mathbf{h}_{\texttt{t-h}}, \mathbf{h}_{\texttt{h-t}}, \mathbf{h}_{\texttt{t-t}}, \mathbf{h}_{\texttt{h-h}})$ is a set of learnable parameters for four meta-relations. 




\subsubsection{Entity Feature Representation}

Similar to relation feature representation, we use entities' connected relation features to represent unseen entities. Considering the direction of relations connected to a specific unseen entity $e$, we calculate the entity feature representation as:
\begin{equation}
    \mathbf{h}_{e} = \frac{1}{|\mathcal{N}(e)|} \sum_{r \in \mathcal{N}(e)} \mathbf{W}^{\text{ent}}_{\text{dir}(r)} \mathbf{h}_{r},
\label{eq:ent-repr}
\end{equation}
where $\mathcal{N}(e)$ is a set of relations connected to $e$; $\mathbf{W}^{\text{ent}}_{\text{dir}(r)}$ 
is a direction-specific parameter for transforming relation features to entity features, and $\mathbf{W}^{\text{ent}}_{\text{dir}(r)} = \mathbf{W}^{\text{ent}}_{\text{in}}$ 
if $r$ is an in-going relation of $e$, or $\mathbf{W}^{\text{ent}}_{\text{dir}(r)} = \mathbf{W}^{\text{ent}}_{\text{out}}$.

\subsubsection{Knowledge Extrapolation with GNN}

We propose a GNN model to encode the entity and relation features.
The previous two modules produce the features for unseen components, and seen components are looked up from learnable feature matrices. 
Our GNN model is based on CompGCN \cite{CompGCN}, which has proven effective in encoding KGs. We extend its entity-relation composition operators to a liner transformation since: 1) this can be viewed as a more general operator between entities and relations; 2) it can use more sophisticated KGE methods as the decoder where the dimension of relations are not equal to that of entities (e.g., RotatE \cite{RotatE}).

For an entity $e$, its message aggregation is conducted as:
\begin{equation}
\begin{aligned}
    \mathbf{m}_{e}^{l+1} = \sum_{(r, t) \in \mathcal{O}(e)} \mathbf{W}_{\text{out}}^l [\mathbf{h}^l_r; \mathbf{h}^l_t] + \sum_{(r,h) \in \mathcal{I}(e)} \mathbf{W}_{\text{in}}^l [\mathbf{h}^l_r; \mathbf{h}^l_h],
\label{eq:gnn-agg}
\end{aligned}
\end{equation}
where $\mathcal{O}(e)$ denotes a set of out-going relations and connected entities, $\mathcal{I}(e)$ denotes a set of in-going relations and connected entities; $\mathbf{W}_{\text{out}}^l$ and $\mathbf{W}_{\text{in}}^l$ are learnable parameters for out- and in-going relation-entity pairs at the $l$-th layer of GNN, $[\cdot;\cdot]$ denotes the concatenation of two vectors, and $\mathbf{h}_r^0 = \mathbf{h}_r$ and $\mathbf{h}_t^0 = \mathbf{h}_t$ are input relation and entity features. After message aggregation from neighbors for each entity, the entity representation for each $e$ is updated by:
\begin{equation}
    \mathbf{h}_{e}^{l+1} = \sigma \left( \frac{\mathbf{m}_{e}^{l+1}}{|\mathcal{O}(e)|+|\mathcal{I}(e)|} + \mathbf{W}_{\text{self}}^{l} \mathbf{h}_{e}^{l} \right),
\label{eq:gnn-update}
\end{equation}
where $\mathbf{W}_{\text{self}}^l$ is a learnable parameter for self-loop updating for entities, and $\sigma$ is an activation function. Besides entities, relations are also updated at each layer:
\begin{equation}
    \mathbf{h}_{r}^{l+1} = \sigma \left( \mathbf{W}_{\text{rel}}^{l} \mathbf{h}_{r}^{l} \right).
\label{eq:gnn-rel-update}
\end{equation}
After updating entity and relation representations, the GNN outputs embeddings for both seen and unseen components in current task, and achieving knowledge extrapolation.

\subsection{Model Learning}

For the task $\mathcal{S}_i$, after obtaining embeddings via previous modules on support triples $\mathcal{T}_{sup}^{i}$, we train our model to score ground-truth query triples $\mathcal{T}_{que}^{i}$ higher than sampled negative triples. 
The score function can be chosen from various KGE methods, and score functions for representative KGE methods can be found in Appendix \ref{app:sec:score-func}.
Furthermore, following previous practice, we use self-adversarial negative sampling loss proposed by~\cite{RotatE} to optimize our model:
\begin{equation}
\begin{aligned}
    \mathcal{L}(\mathcal{S}^{i})= &
    \frac{1}{|\mathcal{T}_{que}^{i}|}
    \sum_{(h,r,t) \in \mathcal{T}_{que}^{i}}
    - \log \sigma\left(
        \gamma + s(h,r,t)
    \right) \\
    &-\sum_{i=1}^{n} p\left(h_{i}^{\prime}, r, t_{i}^{\prime}\right) \log \sigma\left(
        -\gamma-s(h_{i}^{\prime},r,t_{i}^{\prime})
    \right),
\label{eq:self-adv-loss}
\end{aligned}
\end{equation}
where $s(h,r,t)$ is the score for $(h,r,t)$ using embeddings from our model based on $\mathcal{T}_{sup}^{i}$, $\gamma$ is a fixed margin, $n$ is the number of negative samples, $(h_{i}^{\prime},r,t_{i}^{\prime})$ is a negative samples by corrupting a head or tail entity.
$p\left(h_{i}^{\prime}, r, t_{i}^{\prime}\right)$ is the weight for a negative sample, and we put its calculation in Appendix \ref{app:sec:model-learn}.
Finally, we meta-train the model with the overall loss $\sum_{i}\mathcal{L}(\mathcal{S}^i)$ among all tasks sampled from $\mathcal{G}^{tr}$.

\section{Experiments}

In this section, we evaluate our proposed method \model~(for \underline{M}et\underline{a}-Learning Based \underline{K}nowledge \underline{E}xt\underline{r}apolation) on datasets derived from KG benchmarks, and compared it with baselines to show the effectiveness of this model.

\begin{table*}[t]
\centering
\resizebox{\textwidth}{!}{
\begin{tabular}{lcccccc|cccccc}
\toprule
& \multicolumn{6}{c}{FB-Ext} & \multicolumn{6}{c}{NELL-Ext} \\ 
\cmidrule(lr){2-7} \cmidrule(lr){8-13} 
& \multicolumn{2}{c}{\textit{u\_ent}} & \multicolumn{2}{c}{\textit{u\_rel}} & \multicolumn{2}{c}{\textit{u\_both}} & \multicolumn{2}{c}{\textit{u\_ent}} & \multicolumn{2}{c}{\textit{u\_rel}} & \multicolumn{2}{c}{\textit{u\_both}} \\
\cmidrule(lr){2-3} \cmidrule(lr){4-5} \cmidrule(lr){6-7}
\cmidrule(lr){8-9} \cmidrule(lr){10-11} \cmidrule(lr){12-13}
& \multicolumn{1}{c}{MRR} & \multicolumn{1}{c}{Hits@10} & \multicolumn{1}{c}{MRR} & \multicolumn{1}{c}{Hits@10} &
\multicolumn{1}{c}{MRR} & \multicolumn{1}{c}{Hits@10} & \multicolumn{1}{c}{MRR} & \multicolumn{1}{c}{Hits@10} &
\multicolumn{1}{c}{MRR} & \multicolumn{1}{c}{Hits@10} &
\multicolumn{1}{c}{MRR} & \multicolumn{1}{c}{Hits@10} \\
\midrule
GraIL 
& \underline{56.07} & \underline{83.34} & --- & --- & --- & --- 
& \underline{71.62} & \underline{92.92} & --- & --- & --- & --- \\
INDIGO 
& 42.98 & 60.25 & --- & --- & --- & --- 
& 50.31 & 67.68 & --- & --- & --- & --- \\
\midrule
Asmp-KGE (TransE) 
& \underline{63.91} & \underline{82.22} & 33.79 & 35.50 & \underline{13.29} & 23.88 
& \underline{68.64} & \underline{78.35} & 5.26 & 2.50 & 9.25 & 9.65 \\
Asmp-KGE (DistMult) 
& 48.29 & 72.34 & 17.94 & \underline{36.50} & 12.09 & \underline{25.24} 
& 54.00 & 72.18 & 12.77 & 21.67 & 8.05 & 14.35 \\
Asmp-KGE (ComplEx) 
& 46.96 & 67.33 & 11.92 & 22.00 & 12.30 & 24.31
& 51.97 & 67.12 & \underline{14.32} & \underline{22.50} & \underline{9.94} & \underline{17.48} \\
Asmp-KGE (RotatE) 
& 55.12 & 71.65 & \textbf{\underline{34.61}} & 35.50 & 12.96 & 23.42 
& 58.70 & 61.42 & 6.34 & 3.34 & 6.73 & 4.00 \\
\midrule
\model~(TransE) 
& 73.40 & 95.17 & 29.92 & 43.50 & 22.39 & 41.95 
& 70.82 & 92.00 & 24.56 & 54.17 & 21.53 & 51.74 \\
\model~(DistMult) 
& 67.81 & 92.82 & 22.21 & 37.50 & 22.17 & 44.88 
& 70.63 & 91.33 & 27.02 & \textbf{\underline{60.00}} & \textbf{\underline{41.39}} & 57.65 \\
\model~(ComplEx) 
& 70.09 & 93.67 & 24.39 & 43.00 & 24.56 & \textbf{\underline{52.09}}
& 72.24 & 91.91 & 18.27 & 34.17 & 29.39 & 59.65 \\
\model~(RotatE) 
& \textbf{\underline{74.64}} & \textbf{\underline{95.28}} & \underline{32.00} & \textbf{\underline{50.00}} & \textbf{\underline{27.26}} & 49.51  
& \textbf{\underline{77.09}} & \textbf{\underline{94.64}} & \textbf{\underline{31.53}} & 55.00 & 31.45 & \textbf{\underline{62.35}} \\
\bottomrule
\end{tabular}
}
\caption{Link prediction results (\%) on two datasets. We show results for query triples only containing unseen entities (\textit{u\_ent}), only containing unseen relations (\textit{u\_rel}), and containing both unseen entities and relations (\textit{u\_both}). \textbf{Bold} numbers denote the best results and \underline{underline} numbers denote the best results in different kinds of methods. KGE methods after \model~denote the score functions used in \model.}
\label{tab:main-results}
\end{table*}

\subsection{Experimental Setting}

\paragraph{Datasets.}

\begin{table}[ht]
\renewcommand\arraystretch{1.1}
\centering
\resizebox{\columnwidth}{!}{
\begin{tabular}{lccc|cccc}
\toprule
& \multicolumn{3}{c}{Training KG $\mathcal{G}^{tr}$} & \multicolumn{4}{c}{Test KG $\mathcal{G}^{te}$} \\
\cmidrule(lr){2-4} \cmidrule(lr){5-8}
& $|\mathcal{E}^{tr}|$ & $|\mathcal{R}^{tr}|$ & \multicolumn{1}{r}{$|\mathcal{T}^{tr}|$} & $|\mathcal{E}^{te}|$ & $|\mathcal{R}^{te}|$ & $|\mathcal{T}^{te}_{sup}|$ & $|\mathcal{T}^{te}_{que}|$ \\
\midrule
FB-Ext & 952 & 154 & 7,105 & 913 (806) & 196 (56) & 6,103 & 3,524 \\
NELL-Ext & 1,583 & 153 & 5,269 & 851 (753) & 140 (30) & 2,160 & 692 \\
\bottomrule
\end{tabular}
}
\caption{Statistics of the datasets. The number in the bracket denotes the number of entities or relations that doesn't appear in corresponding training KG (i.e., unseen entities or relations).}
\label{tab:data-statistics}
\end{table}

In conventional KG datasets, all entities and relations in test triples are seen during training. 
In order to evaluate the ability of a model for knowledge extrapolation in the federated setting, we create two datasets from two standard KG benchmarks, FB15k-237 \cite{fb15k237} and NELL-995 \cite{DeepPath-NELL995}, named FB-Ext and NELL-Ext. 
For each dataset, we create a training KG $\mathcal{G}^{tr}$ and a test KG $\mathcal{G}^{ts}$ sampled separately from the original benchmark, and a part of entities and relations in the test KG are unseen in the training KG. 
We only put triples that contain at least one unseen component into query triples of the test KG. We also divide query triples into triples only containing unseen entities (\textit{u\_ent}), only containing unseen relations (\textit{u\_rel}), and containing both unseen entities and unseen relations (\textit{u\_both}). 
The statistics of two datasets are given in Table~\ref{tab:data-statistics}. 
The numbers of query triples for \textit{u\_ent}, \textit{u\_rel} and \textit{u\_both} are 1926, 20, 1578 in FB-Ext, and 565, 12, 115 in NELL-Ext. The details of generating datasets can be found in Appendix \ref{app:sec:datasets}.

\paragraph{Baselines.}

We compare our model with two state-of-the-art knowledge graph completion methods considering inductive settings for KGs, GraIL \cite{GraIL}, and INDIGO \cite{INDIGO}, which can handle unseen entities in the test KG. Even though they don't consider unseen relations, they are representative baselines fitting our problems to the best of our knowledge. 
We also compare against baselines that use KGE methods directly on the test KG. Specifically, we first train a KGE model on the training KG. For unseen components in the test KG, we use the corresponding assumption from the score function in the KGE method to calculate their embeddings based on seen components' embeddings.
For example, based on TransE, the embedding for an unseen entity $t$ can be calculated by $\mathbf{h}+\mathbf{r}=\mathbf{t}$ if $h$ and $r$ are seen components and $(h,r,t)$ exists in support triples.
We refer to this baseline as \textit{Asmp-KGE}, and the details are given in Appendix~\ref{app:sec:asmp-kge}.

\paragraph{Evaluation Metrics.}
We report Mean Reciprocal Rank (MRR) and Hits at N (Hits@N) to evaluate the link prediction performance of query triples in the test KG for each dataset. The evaluations consider both head and tail prediction. 
For a fair comparison with baselines, following their settings~\cite{GraIL,INDIGO}, all results are approximated five times by ranking each query triple among 50 other randomly sampled candidate negative triples. 

\paragraph{Implementation Details.}

Our model is implemented in PyTorch and DGL. 
For GraIL and INDIGO, we use the implementations publicly provided by the authors with their best configurations. 
For Asmp-KGE, the dimension is 32.
For \model, the dimensions for embeddings and feature representations are 32; we employ the GNN with 2 layers, and the dimension for GNN's hidden representation is 32. The batch size for meta-training is 64, and we use the Adam optimizer with a learning rate of 0.001.
Before meta-training our model, we sample 10,000 tasks on the training KG for each dataset, and the details of task sampling can be found in Appendix~\ref{app:sec:task-sample}.
During training, we randomly treat entities and relations as unseen with the ratio of 30\% $\sim$ 80\% for each task.

\subsection{Main Results}

We report the link prediction results in Table~\ref{tab:main-results}, and show the detail results for different kinds of query triples (i.e., \textit{u\_ent}, \textit{u\_rel} and \textit{u\_both}) respectively. For GraIL and INDIGO, they can only handle unseen entities, so we leave results for \textit{u\_rel} and \textit{u\_both} blank.
The results show that our proposed \model~achieves improvements over various baselines and has stable performance using different KGE methods.
Most best results are given by \model~with RotatE, a sophisticated KGE model proposed in recent years, showing that our proposed \model~can output reasonable embeddings and make full use of various KGE methods.
Specifically, for \textit{u\_ent} test triples, compared with methods for KG inductive settings (i.e., GraIL and INDIGO), \model~averagely increases by 20.4\% and 8.1\% on MRR and Hits@10 among two datasets, and the numbers are 14.5\% and 18.3\% compared with Asmp-KGE. 
Moreover, by comparing the overall results for different kinds of test triples, we find that the performance for \textit{u\_rel} and \textit{u\_both} triples are worse than the performance for \textit{u\_ent}, which indicates that it is challenging to handle unseen relations in the test KG. Despite the difficulties, our proposed \model~obtains significant improvements on \textit{u\_rel} and \textit{u\_both} triples compared to baselines.
More precisely, \model~averagely increases by 0.56 and	1.02 times for MRR and Hits@10 on \textit{u\_rel} triples and 2.11 and 1.82 times on \textit{u\_both} triples.
Overall, the results show that our model is meta-trained to obtain the ability to extrapolate knowledge for unseen entities and relations and conduct link predictions for those unseen components.

\subsection{Further Analysis}

\paragraph{Ablation Study.}

We conduct several ablation studies to show the importance of different parts of our proposed model. Specifically, we train our model based on following four ablation settings: removing 1) the meta-learning setting (\textit{-Meta}); 2) the relation feature representation (\textit{-RelFeat}); 3) the entity feature representation (\textit{-EntFeat}); 4) the GNN for knowledge extrapolation (\textit{-GNN}). The details of conducting ablation studies above can be found in Appendix~\ref{app:sec:ablation}.
The results of ablation studies using \model~(TransE) on FB-Ext are shown in Table~\ref{tab:ablation}. 
The results show that all ablation settings cause performance to decrease, indicating the importance of these designs.
Moreover, we observe that the meta-learning setting is essential for model performance, indicating the effectiveness of meta-training our model on sampled tasks that mimic the task on the test KG.
We also find that the performance drops significantly after removing the GNN, which is reasonable since the information provided for unseen components just by relation or entity feature representation is limited.


\begin{table}[t]
\linespread{1.4}
\centering
\resizebox{0.85\columnwidth}{!}{
\begin{tabular}{lccccc}
\toprule
& \model & \textit{-Meta} & \textit{-RelFeat} & \textit{-EntFeat} & \textit{-GNN} \\
\midrule
MRR &\textbf{50.31} & 41.29 & 49.00 & 49.62 & 38.83\\
Hits@1 &\textbf{39.00}& 29.89 & 37.06 & 37.88 & 27.36\\
\bottomrule
\end{tabular}
}
\caption{Ablation study of using \model~(TransE) on FB-Ext.}
\label{tab:ablation}
\end{table}

\begin{figure}[t]
\centering
\includegraphics[scale=0.5]{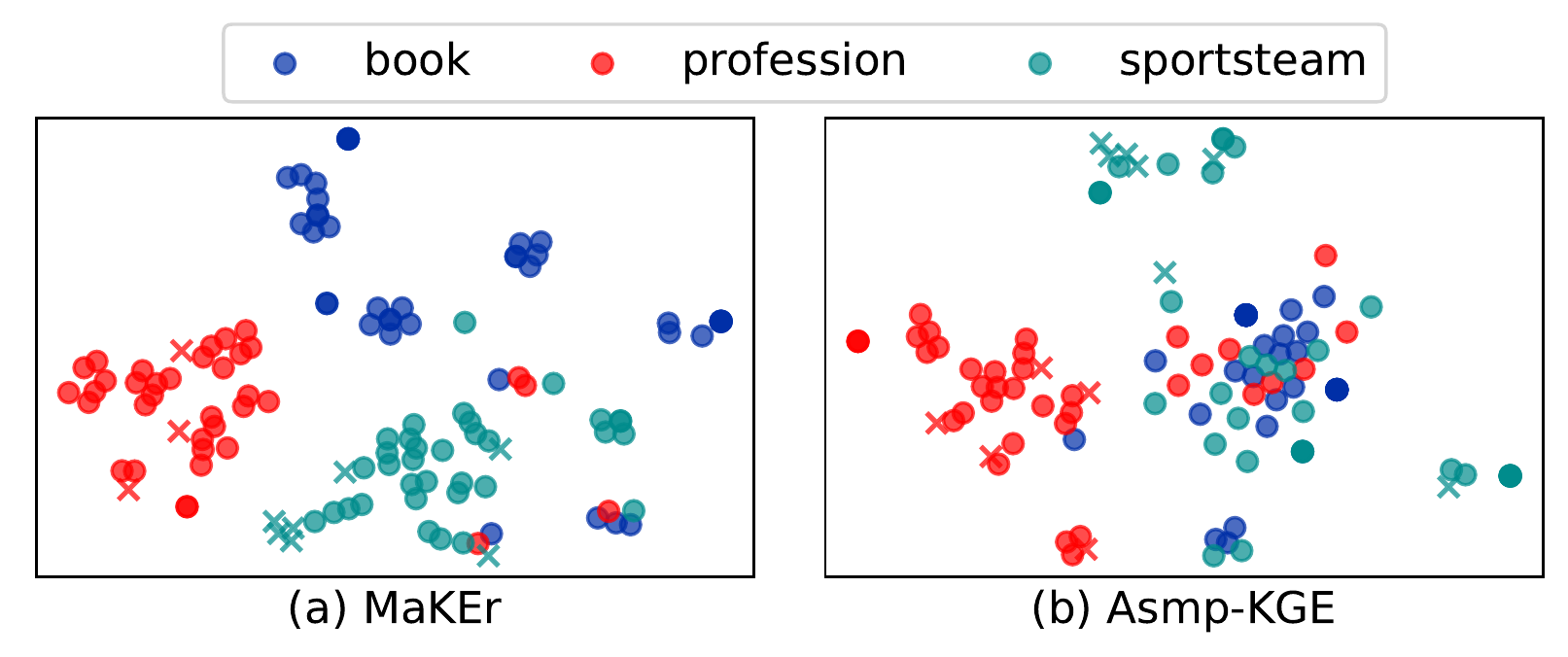}
\caption{Visualization for t-SNE embeddings of \model~(TransE) and Asmp-KGE (TransE). The points using $\circ$ are unseen entities and using $\times$ are seen entities. Their types are marked by colors.}
\label{fig:ent-tsne}
\end{figure}

\paragraph{Case Study for Unseen Entities.}
We visualize the entity embeddings for NELL-Ext produced by our proposed \model~and Asmp-KGE in Fig.~\ref{fig:ent-tsne}. 
In this figure, we show different types of entities with different colors. 
The distributions of embeddings from \model~are more consistent with their corresponding types than the embeddings produced by Asmp-KGE.
For Asmp-KGE, the embeddings of different entity types are mixed, while for \model, the embeddings are mapped into different clusters.
Furthermore, we also find that in our model, the embeddings for unseen entities can be clustered with seen entities in the same type.
The clustering for entities from different types indicates that \model~can represent unseen entities with embeddings containing reasonable semantics and informative knowledge.

\begin{figure}[t]
\centering
\includegraphics[scale=0.41]{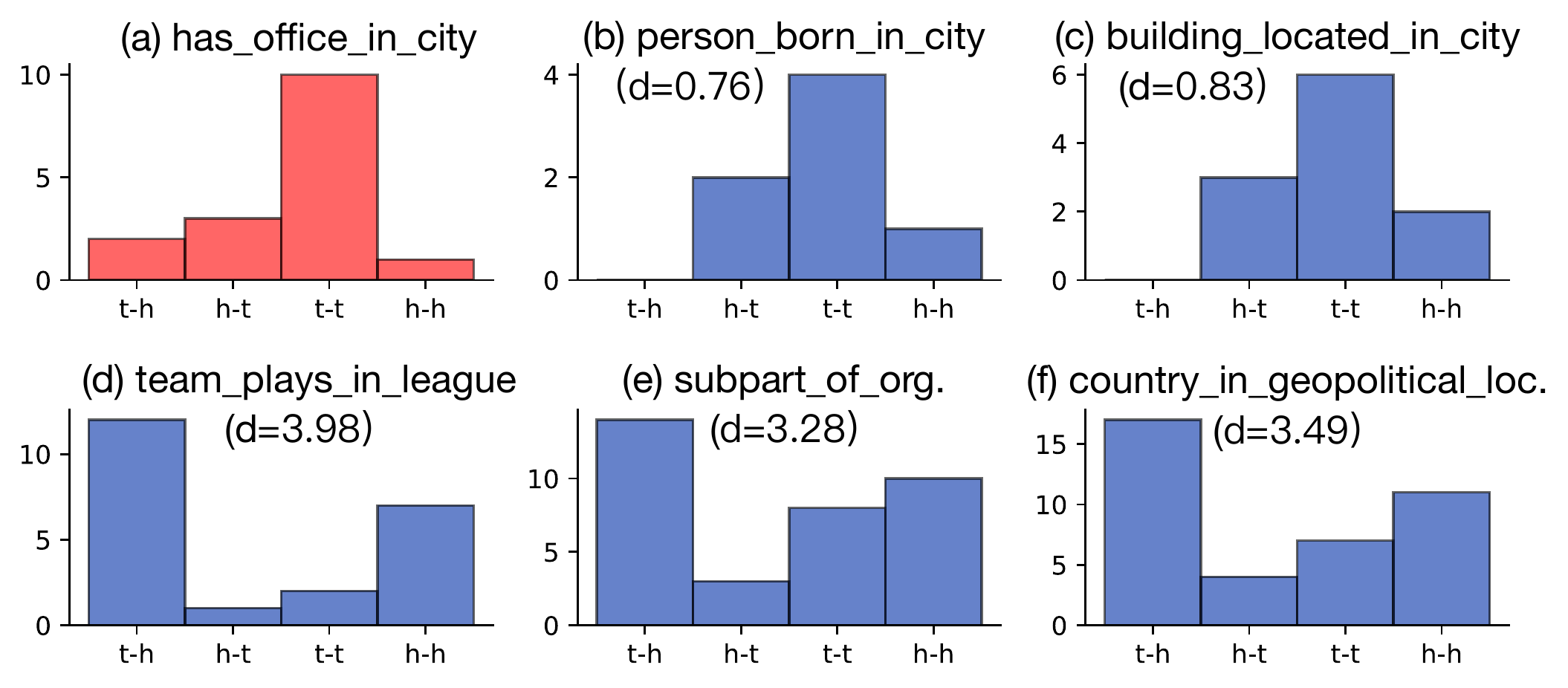}
\caption{The number of connected meta-relations in RPG for different relations. (a) is an unseen relation, and we show the euclidean distances between (a) and other relations in brackets based on embeddings produced by \model~(TransE).}
\label{fig:rel-bar}
\end{figure}

\paragraph{Case Study for Unseen Relations.}
From Fig.~\ref{fig:rel-bar}, we find that for an unseen relation \texttt{has\_office\_in\_city} in (a), its distribution of meta-relations is more similar to relations in (b)(c) which has similar semantics compared to relations in (d)(e)(f). 
Furthermore, three relations in the bottom half all represent the relationships about subordination, and they also have similar meta-relation distributions. These observations show that the connected meta-relations for rel-nodes in RPG are sensitive to the semantics of corresponding relations. We think this is also why such feature representations are effective for relations.
Moreover, from the distances between the unseen relation in (a) and other relations, we find that the embeddings produced by \model~are reasonable where similar relations are close in the vector space, showing the effectiveness for embedding unseen relations in our proposed \model.

\section{Conclusion}

In this paper, we formulate the problem of embedding unseen entities and relations for an emerging KG in the federated setting. 
To solve this problem, we resort to meta-learning and sample tasks on the training KG to mimic the link prediction task in the test KG. 
We design a GNN framework that can represent the features for unseen components and output embeddings for them. This model can generalize to unseen components of the test KG after being meta-trained on a set of tasks with unseen components.
Our future work might consider designing explainable feature representations modules that can induce rules to make our model more human-friendly.

\clearpage

\section*{Acknowledgments}
This work is funded by NSFC U19B2027/91846204.

\bibliographystyle{named}
\bibliography{ijcai22}

\clearpage

\appendix

\section{Score functions For KGE Methods}
\label{app:sec:score-func}

Many score functions of existing KGE methods can be used in our model. In this paper, we choose four representative KGE methods, TransE~\cite{TransE}, DistMult~\cite{DistMult}, ComplEx~\cite{ComplEx} and RotatE~\cite{RotatE}, to evaluate our proposed model. The details of their score functions are shown in Table~\ref{app:tab:score-function}.

\begin{table}[h]
\centering
\begin{tabular}{lcc}
\toprule
Model & Score Function & Vector Space \\
\midrule
TransE & $-|| \mathbf{h} + \mathbf{r} - \mathbf{t} ||$ & $ \mathbf{h}, \mathbf{r}, \mathbf{t} \in \mathbb{R}^d $ \\
DistMult & $\mathbf{h}^\top \operatorname{diag}(\mathbf{r}) \mathbf{t}$ & $ \mathbf{h}, \mathbf{r}, \mathbf{t} \in \mathbb{R}^d $ \\
ComplEx & $\operatorname{Re}(\mathbf{h}^\top \operatorname{diag}(\mathbf{r}) \overline{\mathbf{t}})$ & $ \mathbf{h}, \mathbf{r}, \mathbf{t} \in \mathbb{C}^d $ \\
RotatE & $-|| \mathbf{h} \circ \mathbf{r} - \mathbf{t} ||$ & $ \mathbf{h}, \mathbf{r}, \mathbf{t} \in \mathbb{C}^d $ \\
\bottomrule
\end{tabular}
\caption{Score function $s(h, r, t)$ of typical knowledge graph models. $\mathbf{h}, \mathbf{r}, \mathbf{t}$ are embeddings correspond to $h, r, t$. $\operatorname{Re(\cdot)}$ denotes the real vector component of a complex valued vector. $\circ$ denotes the Hadamard product.}
\label{app:tab:score-function}
\end{table}

\section{Model Learning}
\label{app:sec:model-learn}

We use self-adversarial negative sampling loss proposed by~\cite{RotatE} to optimize our model:
\begin{equation}
\begin{aligned}
    \mathcal{L}(\mathcal{S}^{i})= &
    \frac{1}{|\mathcal{T}_{que}^{i}|}
    \sum_{(h,r,t) \in \mathcal{T}_{que}^{i}}
    - \log \sigma\left(
        \gamma + s(h,r,t)
    \right) \\
    &-\sum_{i=1}^{n} p\left(h_{i}^{\prime}, r, t_{i}^{\prime}\right) \log \sigma\left(
        -\gamma-s(h_{i}^{\prime},r,t_{i}^{\prime})
    \right),
\end{aligned}
\end{equation}
where $p\left(h_{i}^{\prime}, r, t_{i}^{\prime}\right)$ is the self-adversarial weight for a negative triple among a set of negative samples, and it is calculated by:
\begin{equation}
    p\left(h_j^{\prime}, r, t_{j}^{\prime}\right) = \frac{\exp \alpha s(h_j^{\prime}, r, t_{j}^{\prime})}{\sum_{i} \exp \alpha s(h_i^{\prime}, r, t_{i}^{\prime})}, 
\label{eq:adv-sample}
\end{equation}
where $\alpha$ is the temperature of sampling.

\section{Datasets}
\label{app:sec:datasets}

To generate the datasets for the problem of knowledge extrapolation in the federated setting, we first sample a set of entities $\mathcal{E}_{1}$ from the original KG dataset $\mathcal{G}$, and conduct random walk with length $l_1$ from those entities to get an expanded entity set $\mathcal{E}_{1}^{\prime}$. 
Second, we extract triples based on the entities in $\mathcal{E}_{1}^{\prime}$ to get the triples for the test KG, and remove such triples from $\mathcal{G}$. 
To ensure that there are entities and relations in the test KG are unseen in the training KG, we also remove a part of entities and relations from $\mathcal{G}$ with the ratio of $r_1$.
We also extract a validation KG similar to the test KG following previous steps.

After obtaining the test and validation KG, we sample the training KG on the remaining KG $\mathcal{G}^{\prime}$. To extract the training KG, we first sample a set of entities $\mathcal{E}_{2}$ from $\mathcal{G}^{\prime}$, and conduct random walk with length $l_2$ from those entities to get an expanded entity set $\mathcal{E}_{2}^{\prime}$. We extract triples based on the entities in $\mathcal{E}_{2}^{\prime}$ to get the triples for the training KG.
The parameters for sampling datasets are shown in Table~\ref{app:tab:sample-parameter}. 

\begin{table}[ht]
\centering
\begin{tabular}{lll}
\toprule
Parameters & FB-Ext & NELL-Ext \\
\midrule
$|\mathcal{E}_{1}|$ & 100 & 100 \\ 
$|\mathcal{E}_{2}|$ & 100 & 200 \\
$l_{1}$ & 10 & 15 \\ 
$l_{2}$ & 10 & 20 \\ 
$r_{1}$ & 0.1 & 0.1 \\
\bottomrule
\end{tabular}
\caption{Parameters of dataset sampling for two datasets.}
\label{app:tab:sample-parameter}
\end{table}

The detail statistics for validation KGs of two datasets are shown in Table~\ref{app:tab:valid-data-statistics}.

\begin{table}[ht]
\renewcommand\arraystretch{1.1}
\centering
\resizebox{\columnwidth}{!}{
\begin{tabular}{lcccc}
\toprule
& \multicolumn{4}{c}{Validation KG $\mathcal{G}^{va}$} \\
\cmidrule(lr){2-5}
& $|\mathcal{E}^{va}|$ & $|\mathcal{R}^{va}|$ & $|\mathcal{T}^{va}_{sup}|$ & $|\mathcal{T}^{va}_{que}|$ \\
\midrule
FB-Ext & 908 (801) & 174 (42) & 6,687 & 1,672 \\
NELL-Ext & 583 (507) & 109 (20) & 1,242 & 309 \\
\bottomrule
\end{tabular}
}
\caption{Statistics of validation KGs of datasets. The number in the bracket denotes the number of entities or relations that doesn't appear in corresponding training KG (i.e., unseen entities or relations).}
\label{app:tab:valid-data-statistics}
\end{table}

\section{Details of Asmp-KGE}
\label{app:sec:asmp-kge}

The calculations for Asmp-KGE are based on a specific KGE methods, and there are three embedding calculation operations for each KGE method: 1) $f_\text{hr2t}$, calculating the tail embedding based on the head and relation embeddings; 2) $f_\text{tr2h}$ calculating the head embedding based on the tail and relation embeddings; 3) $f_\text{ht2r}$ calculating the relation embedding based on the head and tail embeddings.

For an unseen entity $e$ in the test KG $\mathcal{G}^{te}$, based on $\mathcal{T}^{te}_{sup}$, we first find all triples related to $e$ that other two components are a seen entity and relation, then we use $f_\text{hr2t}$ or $f_\text{tr2h}$ to calculate its embedding for each related triples and finally take an average. For an unseen relation, we use the similar steps to get its embedding. For unseen entities and relations which have no related triples that other two components are seen components, we use the average embeddings of all entities and relations as the embeddings for such unseen entities and relations. We show calculations for different KGE methods in Asmp-KGE as follows, and we use $\textbf{h}$, $\textbf{r}$ and $\textbf{t}$ to denote the embeddings for a triples $(h,r,t)$.

TransE:
\begin{equation}
\begin{aligned}
    f_\text{hr2t}(\textbf{h}, \textbf{r}) &= \textbf{h} + \textbf{r}, \\
    f_\text{tr2h}(\textbf{t}, \textbf{r}) &= \textbf{t} - \textbf{r}, \\
    f_\text{ht2r}(\textbf{h}, \textbf{t}) &= \textbf{t} - \textbf{h}.
\end{aligned}
\end{equation}

DistMult:
\begin{equation}
\begin{aligned}
    f_\text{hr2t}(\textbf{h}, \textbf{r}) &= \textbf{h} \circ \textbf{r}, \\
    f_\text{tr2h}(\textbf{t}, \textbf{r}) &= \textbf{t} \circ \textbf{r}, \\
    f_\text{ht2r}(\textbf{h}, \textbf{t}) &= \textbf{h} \circ \textbf{t}.
\end{aligned}
\end{equation}

ComplEx:
\begin{equation}
\begin{aligned}
    f_\text{hr2t}(\textbf{h}, \textbf{r}) &= \textbf{h} \circ \textbf{r}, \\
    f_\text{tr2h}(\textbf{t}, \textbf{r}) &= \overline{\overline{\textbf{t}} \circ \textbf{r}}, \\
    f_\text{ht2r}(\textbf{h}, \textbf{t}) &= \overline{\textbf{h} \circ \overline{\textbf{t}}}.
\end{aligned}
\end{equation}

RotatE:
\begin{equation}
\begin{aligned}
    f_\text{hr2t}(\textbf{h}, \textbf{r}) &= \textbf{h} \circ \textbf{r}, \\
    f_\text{tr2h}(\textbf{t}, \textbf{r}) &= \textbf{t} / \textbf{r}, \\
    f_\text{ht2r}(\textbf{h}, \textbf{t}) &= \textbf{t} / \textbf{h}.
\end{aligned}
\end{equation}

\section{Sampling for Meta-Learning Tasks}
\label{app:sec:task-sample}

The sampling procedure is as follows: 
1) first, we randomly choose an entity and conduct a random walk from it to form a set of selected entities; 2) then, we choose the next entity from the above entity set and conduct a random walk to expand the entity set; 
3) we conduct the last step multiple times and use the triples consisting of entities from the final entity set as sampled triples; 
4) the sampled triples are randomly split into support and query triples for the current task. 

\section{Details of Ablation Study}
\label{app:sec:ablation}

The details of conducting ablation studies are described as follows.

\begin{enumerate}
    \item For removing meta-learning (\textit{-Meta}), we don't meta-train our model based on a set of sampled tasks described in Sec.~\ref{sec:meta-learning-setting}, but directly train it on all triples in the training KG and use the same triples to calculate the loss function. The training procedure is the same as conventional KGE methods.
    \item For ablating relation feature representation (\textit{-RelFeat}), this module is replaced by using random vector representations to represent unseen relations. Furthermore, to make random representation as realistic as possible, we use the maximum and minimum values of seen relations' features to limit the random feature generation.
    \item For ablating entity feature representation (\textit{-EntFeat}), this module is replaced by using random vector representations to represent unseen entities, and we use the same way as ablating relation feature representation to bound the random feature generation.
    \item For removing knowledge extrapolation with GNN (\textit{-GNN}), we directly treat the feature representations for entities and relations as their embeddings.
\end{enumerate}

\end{document}